\documentclass{article}

\PassOptionsToPackage{numbers, compress}{natbib}

\usepackage[preprint]{neurips_2019}

\usepackage[utf8]{inputenc} 
\usepackage[T1]{fontenc}    
\usepackage{hyperref}       
\usepackage{url}            
\usepackage{booktabs}       
\usepackage{amsfonts}       
\usepackage{nicefrac}       
\usepackage{microtype}      
\usepackage{amsmath}

\usepackage{color}
\usepackage{ulem}
\usepackage{graphicx}
\usepackage{subcaption}
\usepackage{lipsum}

\newcommand{\Skip}[1]{}

\newcommand{\eg}{e.g., }

\newcommand{\figref}[1]{Figure~\ref{#1}}

\title{IKEA Furniture Assembly Environment\\ for Long-Horizon Complex Manipulation Tasks}
\author{%
  Youngwoon Lee, Edward S. Hu, Zhengyu Yang, Alex Yin, and Joseph J. Lim\\
  Department of Computer Science\\
  University of Southern California\\
  \texttt{\{ lee504, hues, yang765, alexyin, limjj \}@usc.edu} \\
  \color{red}{\url{https://clvrai.com/furniture}} \\
}

\begin{document}

\maketitle
\begin{abstract}
The \textit{IKEA Furniture Assembly Environment} is one of the first benchmarks for testing and accelerating the automation of complex manipulation tasks. The environment is designed to advance reinforcement learning from simple toy tasks to complex tasks requiring both long-term planning and sophisticated low-level control. Our environment supports over 80 different furniture models, Sawyer and Baxter robot simulation, and domain randomization. The IKEA Furniture Assembly Environment is a testbed for methods aiming to solve complex manipulation tasks. The environment
is publicly available at \color{red}{\url{https://clvrai.com/furniture}}.
\end{abstract}

\begin{figure*}[ht]
\centering
    \includegraphics[width=.8\textwidth]{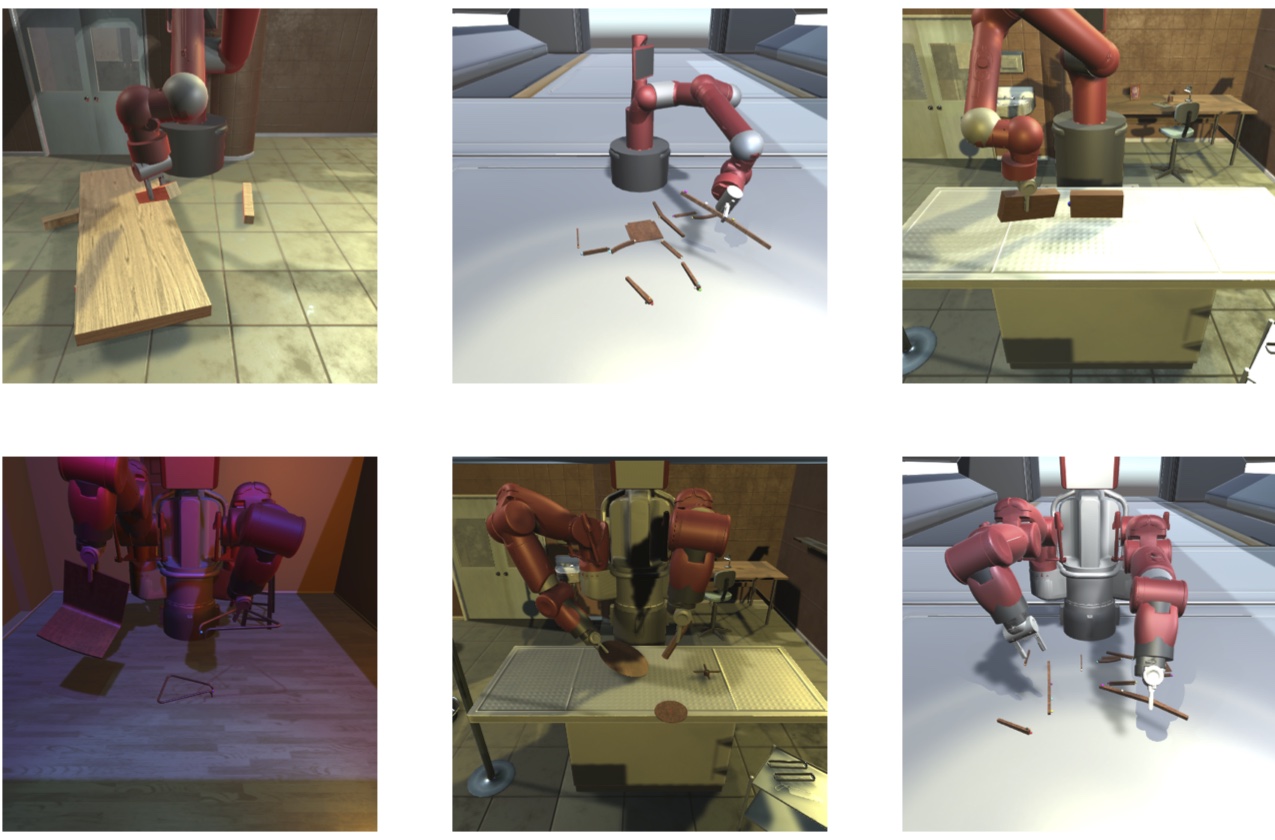}
    \caption{
        The \textit{IKEA Furniture Assembly Environment} is a furniture assembly simulator. It contains diverse sets of furniture models and robots, and supports various background, lighting, and textures.
    }
    \label{fig:teaser}
\end{figure*}

\section{Introduction}
The ability to plan and manipulate physical objects is a necessity to use tools, build structures, and ultimately interact with the world in a meaningful way. Thus, solving long-horizon manipulation tasks has been an active challenge in robot learning. One main bottleneck is the lack of a "standardized" simulation environment for long-term physical tasks. Such a simulator for robot learning needs to have the following properties: long-horizon hierarchical tasks for planning, high-quality rendering for sim-to-real transfer, variable agents and dynamic domains.

Imagine that you are building IKEA furniture. It is not trivial to figure out how to assemble pieces into the final configuration given only a glimpse of fully constructed furniture. Specifically, it is not apparent from pieces on the floor which parts to choose for attachment and in what order. Hence, we need to dissect the final configuration and deduce the sequence of tasks necessary to build the furniture by comparing the current and final configuration. Moreover, connecting two parts requires complicated manipulation skills, such as accurate alignment of two attaching points and sophisticated force control to firmly attach them. Therefore, furniture assembly is a comprehensive task requiring reliable perception, high-level planning, and sophisticated control, making it a suitable benchmark for robot learning algorithms. 

To this end, we introduce the \textit{IKEA Furniture Assembly Environment} as a new benchmark for complex autonomous manipulation skills. The IKEA Furniture Assembly Environment is a visually realistic environment that simulates the task of furniture assembly as a step toward autonomous manipulation. The furniture assembly task involves not only high-level 3D scene understanding and step-by-step planning but also sophisticated low-level control. Figure \ref{fig:teaser} shows examples of our rendered environments.

A variety of research problems could be investigated with this new environment, namely perception, planning, and control. For perception, the environment could be used to solve 3D object detection, pose estimation, instance segmentation, scene graph generation, and shape estimation problems. For robotic control, the environment is suitable for testing multi-agent reinforcement learning, hierarchical reinforcement learning, model-based reinforcement learning, imitation learning, and sim-to-real algorithms for long-term complex manipulation tasks. 

The IKEA Furniture Assembly Environment simulates over 80 furniture models. The environment supports multiple agents such as Cursor, Baxter and Sawyer robots. To test generalization, the environment supports domain randomization in the furniture, physics, lighting, textures, and more factors of variation found in the real world. We also outline in the conclusion promising future directions for complex manipulation tasks.

\begin{figure}[t]
    \centering
    \begin{subfigure}[t]{0.222\textwidth}
    	\centering
    	\includegraphics[width=\textwidth]{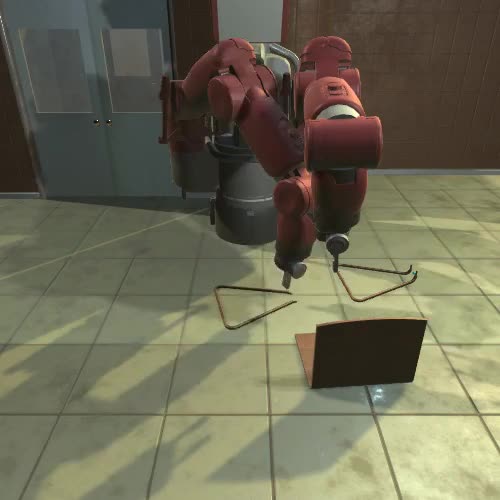}
    	\caption{Initialization}
    \end{subfigure}
    \quad
    \begin{subfigure}[t]{0.222\textwidth}
    	\centering
    	\includegraphics[width=\textwidth]{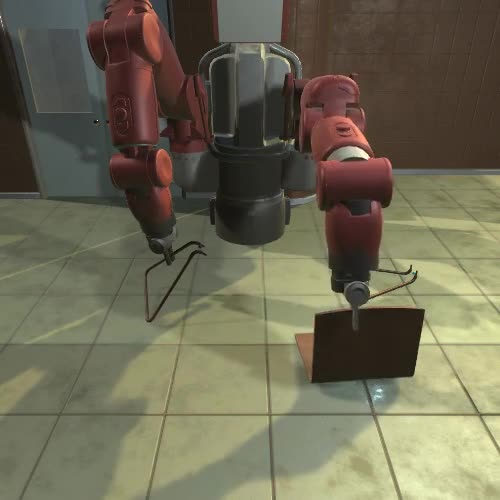}
    	\caption{Grasping 2 parts}
    \end{subfigure}
    \quad
    \begin{subfigure}[t]{0.222\textwidth}
    	\centering
    	\includegraphics[width=\textwidth]{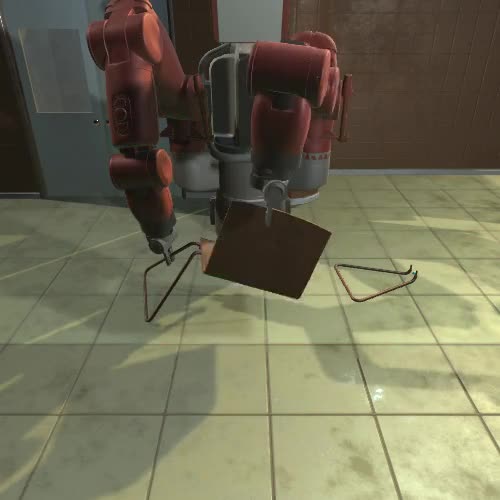}
    	\caption{Aligning 2 parts}
    \end{subfigure}
    \quad
    \begin{subfigure}[t]{0.222\textwidth}
    	\centering
    	\includegraphics[width=\textwidth]{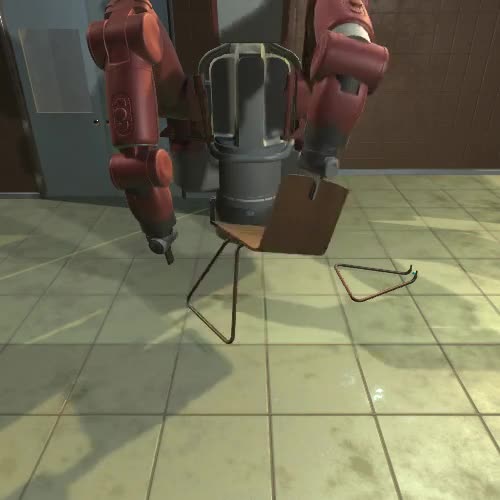}
    	\caption{Attaching 2 parts}
    \end{subfigure}
    \caption{
        The proposed IKEA Furniture Assembly Environment simulates robotic furniture assembly. The robot decides which parts to attach (a), grasps the desired parts (b), aligns the grasped parts (c), and then attaches the grasped parts (d).
    }
    \label{fig:4stepassembly}
\end{figure}

\begin{figure}[t]
    \centering
    \begin{subfigure}[t]{0.242\textwidth}
    	\centering
    	\includegraphics[width=\textwidth]{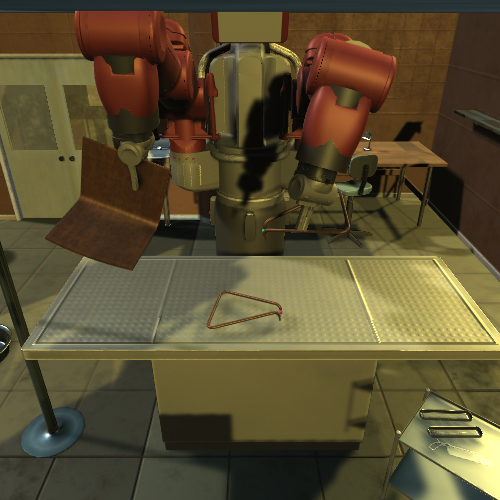}
        \caption{RGB}\label{fig:camera_ob}
    \end{subfigure}
    \begin{subfigure}[t]{0.242\textwidth}
    	\centering
    	\includegraphics[width=\textwidth]{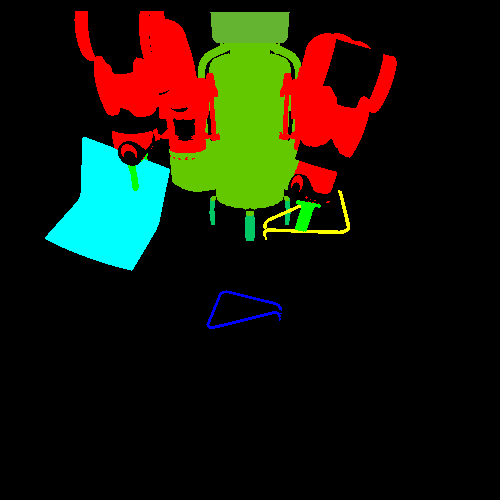}
        \caption{Segmentation}\label{fig:segmentation_ob}
    \end{subfigure}
    \begin{subfigure}[t]{0.242\textwidth}
    	\centering
    	\includegraphics[width=\textwidth]{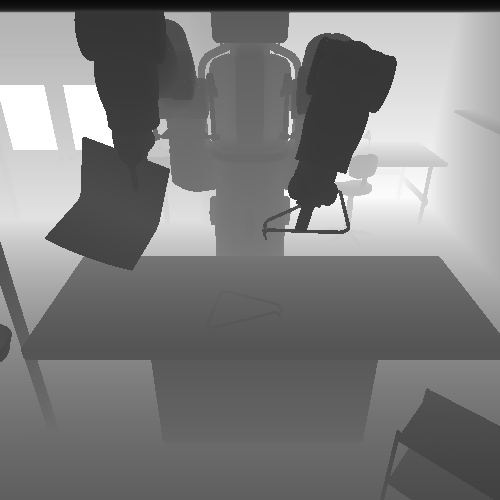}
        \caption{Depth}\label{fig:depth_ob}
    \end{subfigure}
    \begin{subfigure}[t]{0.242\textwidth}
    	\centering
    	\includegraphics[width=\textwidth]{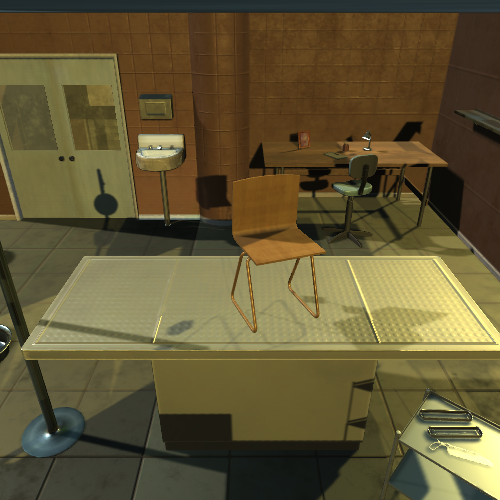}
        \caption{Goal}\label{fig:goal_ob}
    \end{subfigure}
    \caption{
        An RGB image, part segmentation, depth map, goal image, and agent state (joint positions and velocities) are available observations. The agent can interact with the environment by either generating torque for each joint or choosing a motion primitive to attach the two parts by aligning the connectors.
    }
    \label{fig:observations}
\end{figure}

\section{IKEA Furniture Assembly Environment}
To advance reinforcement learning from simple, videogame-esque tasks to complex and realistic tasks, our environment features long-horizon, hierarchical tasks, realistic rendering, domain randomization, and accurate robotic simulation. 
The furniture assembly task is composed of (1) selecting two compatible parts, (2) grasping these two parts, (3) aligning the attachable spots (namely connectors) of the parts, and (4) connecting the parts as seen in \figref{fig:4stepassembly}. The furniture assembly task can be accomplished by repeating this process until all parts are assembled.
We first discuss various research topics that can utilize our environment, and then present the design and implementation of the environment.

\subsection{Target Research Topics}
The IKEA Furniture Assembly environment simulates complex long-term robotic manipulation tasks with 80+ furniture models and different robots. Moreover, the environment features realistic rendering, configurable scene (\eg lighting, texture, color), and diverse ground truth labels, such as object pose, shape, instance segmentation mask, depth map, and scene graph.
Hence, the environment can be used not only for learning planning and control but also for learning perception as following:

\begin{itemize}
    \item \textbf{Computer Vision:} The environment can generate an abundant amount of data with diverse labels, such as object pose, shape, instance segmentation mask, depth map, part configurations, and scene graph. This synthetic data can be used to tackle many computer vision problems, such as object pose estimation, semantic segmentation, scene graph understanding, shape estimation.

    \item \textbf{Control:} The furniture assembly task requires sophisticated manipulation skills. We support impedance control, inverse kinematics for task space control, as well as keyboard control for humans. Thus, the proposed environment can be used as a challenging benchmark for reinforcement learning and imitation learning methods. Bi-manual or multi-agent manipulation of furniture is another interesting direction.

    \item \textbf{Planning:} The furniture assembly consists of multiple steps of assembling two furniture parts and thus has a long horizon. To tackle this long-horizon task, an agent should learn to plan using approaches from model-based planning and hierarchical reinforcement learning.

    \item \textbf{Others:} The visual and interaction data can be collected to acquire domain knowledge, such as intuitive physics, for other applications. Language-guided robot learning~\citep{oh2017zero-shot,chaplot2017gated} is also a promising research direction to solve complex tasks.
\end{itemize}


\subsection{Environment Development}
To cover several challenges including 3D alignment of various shapes of objects and long-horizon robotic manipulation, we developed a novel 3D environment that supports assembling IKEA furniture models using MuJoCo~\citep{todorov2012mujoco} as a physics simulator and Unity3D game engine as a renderer. MuJoCo provides fast and accurate physics simulation, while Unity3d has superior texture and lighting configuration. To exploit the strengths of both frameworks, we use MuJoCo as the underlying physics engine and Unity3d as the renderer. To enable the robotic simulation of Sawyer and Baxter arms in MuJoCo, we use the Robosuite framework~\citep{fan2018surreal}. We also use DoorGym~\citep{urakami2019doorgym} to integrate Unity rendering engine with MuJoCo.

\subsection{Assembly Simulation}
The simulation environment follows the OpenAI Gym~\citep{brockman2016openai} protocol where an environment takes an action as input and takes a step. 
Robotic arms can move around the environment and interact with furniture parts.
In addition to actions for robotic arm moves, our environment has a \textit{connect} action.
During each step, the environment checks all pairs of connectors. Currently, we only support one-to-one mapping between connectors and plan to implement many-to-many mappings for identical parts. If a pair of connectors have matching IDs and the connectors are within a positional and angular distance threshold, the pair of connectors are flagged as attachable. Specifically, we check the euclidean distance between the connector coordinates (Equation~\eqref{eq:pos}), the cosine distance between the connector up vectors (Equation~\eqref{eq:up}), and the cosine distance between the connector forward vectors (Equation~\eqref{eq:forward}). 
If the \textit{connect} action is activated, then the attachable parts are connected using the MuJoCo weld mechanism. The thresholds for distance and angle are configurable. For full details, refer to the \texttt{\_is\_aligned} function in the \texttt{furniture.py} file.
\begin{equation}
   Pos = d_\text{L2}((x,y,z)_A, (x,y,z)_B) < \epsilon_\text{distance}
\label{eq:pos}
\end{equation}
\begin{equation}
   Up =  d_\text{cos}(Up_A, Up_B) > \epsilon_\text{up}
\label{eq:up}
\end{equation}
\begin{equation}
   Forward = d_\text{cos}(Forward_A, Forward_B) > \epsilon_\text{forward}
\label{eq:forward}
\end{equation}
\begin{equation}
    Attachable=
    \begin{cases}
      1 & \text{if}\ Pos \land Up \land Forward \\
      0 & \text{otherwise}
    \end{cases}
    \label{eq:attachable}
\end{equation}

\paragraph{Tasks and Reward Functions}
Currently, the environment provides a sparse reward for successfully connecting two furniture parts. Users can easily implement their own reward functions and use the rewards for training an agent. For example, the \textsc{Baxter-Block-Pick} task implemented in the \texttt{furniture\_baxter\_block.py} file computes the reward function composed of 10 different components including the distance between the gripper and the target object, angle of the gripper, height of the target object, and gripper's state. 


\subsection{Furniture Models}
\begin{figure}[t]
\centering
    \includegraphics[width=1\textwidth]{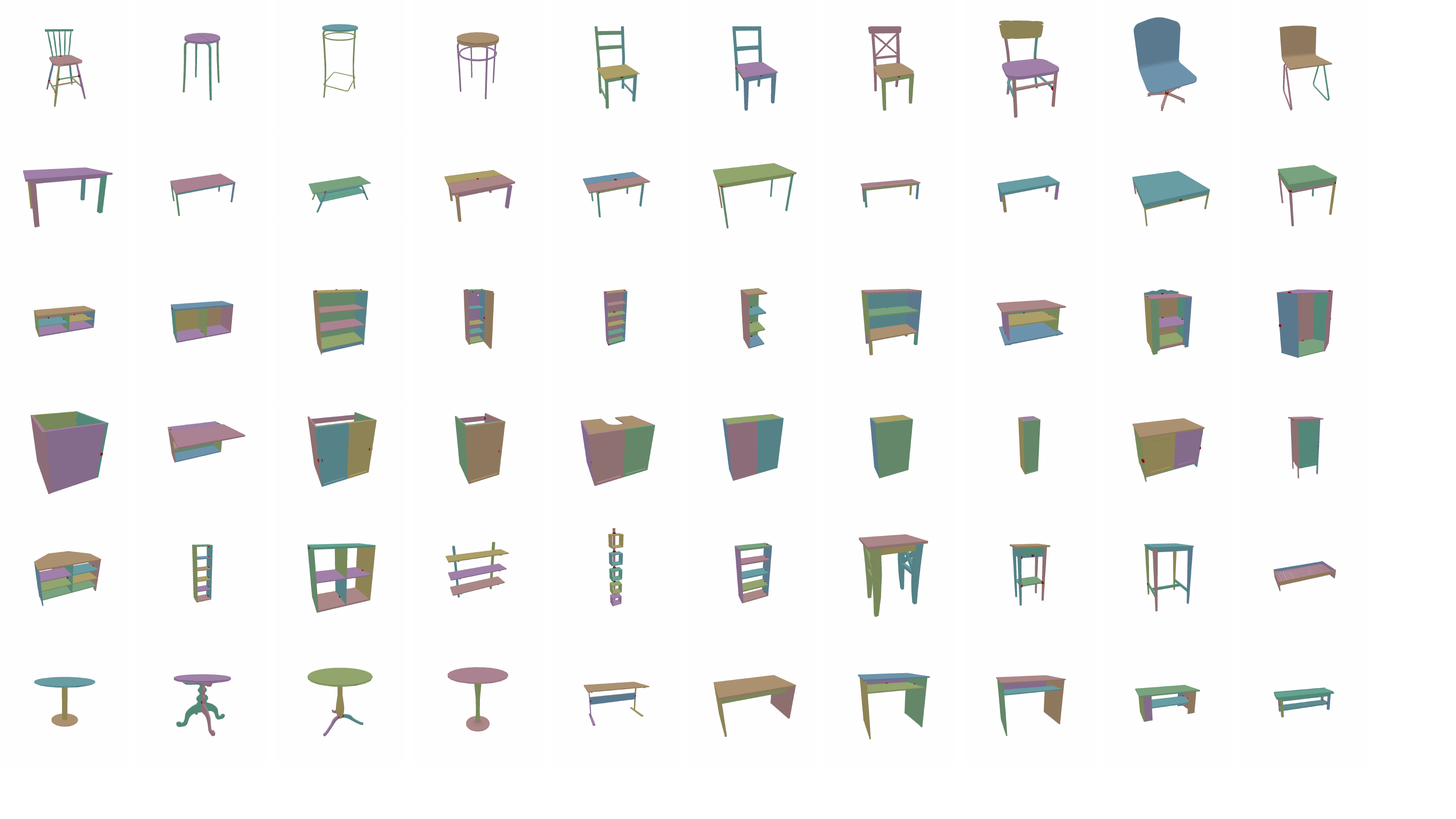}
    \caption{
        The IKEA furniture assembly environment includes diverse set of furniture models. Each furniture is modeled following the IKEA's user's manual and different parts are colored differently.
        \label{fig:models}
    }
\end{figure}

Our environment provides a simulation of the furniture assembly of over 80 different furniture models as shown in \figref{fig:models}. We define the furniture as a combination of parts and connectors.

\paragraph{Parts}
The furniture parts are modeled by following the IKEA's official user's manuals with a minor simplification in small details such as carving and screws. 
The furniture models are created using 3D modeling tool Rhino\footnote{\url{https://www.rhino3d.com/}}
and each furniture part is converted to a separate 3D mesh file in a format of STL.
Given an STL 3D mesh file, MuJoCo can represent the mesh as a physical object. Each furniture part is represented as one or more meshes. Concave furniture parts need to be represented with multiple meshes, as MuJoCo only supports convex meshes for collision detection. To enable collision of a concave mesh, we split a concave mesh into multiple convex meshes with STL editing software or use MuJoCo's primitive meshes (\eg box, cylinder, sphere) only for the collision detection.

\paragraph{Connectors}
A pair of connectors define the connection information between two furniture parts. The connectors are located on the parts and serve as areas of attachment. On a given part, we parameterize connectors with their ID, size, position, and orientation to the part. All furniture parts and connectors are defined through the MuJoCo XML file. Refer to the XML documentation in the codebase for information and examples on the schema.

\paragraph{Assembly Information}
To handle part connection, we define in the XML the connector constraints (e.g. Part A can connect to Part B in a certain pose) with the MuJoCo \textit{weld} equality constraint, which allows for welding together two given parts. Once all parts and their connector constraints are specified, the furniture model is complete and can be loaded into the MuJoCo simulator for simulation.

\subsection{Agents}
\begin{figure}[h]
    \centering
    \begin{subfigure}[t]{0.31\textwidth}
    	\centering
    	\includegraphics[width=\textwidth]{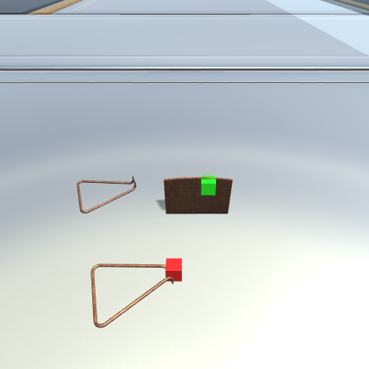}
        \caption{Cursor}\label{fig:cursor}
    \end{subfigure}
    \quad
    \begin{subfigure}[t]{0.31\textwidth}
    	\centering
    	\includegraphics[width=\textwidth]{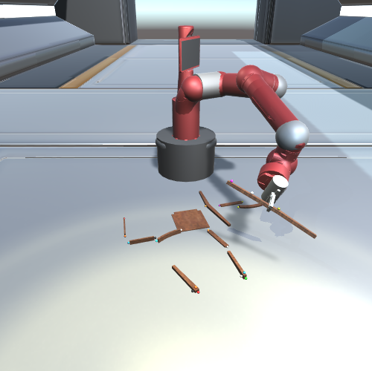}
        \caption{Sawyer}\label{fig:sawyer}
    \end{subfigure}
    \quad
    \begin{subfigure}[t]{0.31\textwidth}
    	\centering
    	\includegraphics[width=\textwidth]{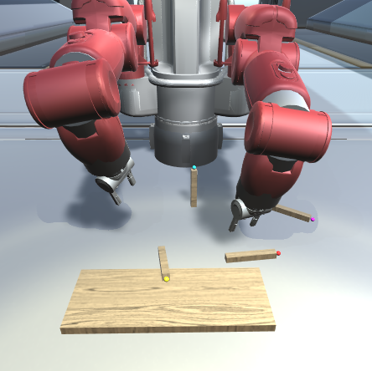}
        \caption{Baxter}\label{fig:baxter}
    \end{subfigure}
    \caption{
        The Cursor, Sawyer, and Baxter agents are supported. We plan to add Fetch, UR, Jaco, and more robots in the next update.
    }
    \label{fig:agents}
\end{figure}

The environment supports a variety of agents for interacting with furniture. Currently, three agents are available: Cursor, Sawyer, and Baxter, as illustrated in \figref{fig:agents}. Their \textit{action spaces} and \textit{observation spaces} are fully configurable to fit a variety of problem settings. The observation space can consist of agent state (\eg joint positions and velocities), environmental state (\eg coordinates and rotation of furniture parts), and a camera observation. Aside from RGB images, the environment also supports object segmentation masks and depth camera observations as seen in \figref{fig:observations}. In general, the action space consists of movement, selection, and attachment primitives.

\paragraph{Cursor}
 The \textit{Cursor} agent abstracts away the low-level object grasping problem. Composed of two floating and collision-free cursors in the environment, the agent can move the cursors in the (x,y,z)-axes, and hold parts that are encapsulated by the cubes. Parts that are held by the cursor can be rotated on the (x,y,z)-axes. Movement and rotation can either be applied continuously or in discrete steps depending on the configuration. This agent is suitable for methods that focus on the planning and reasoning portion of the furniture assembly problem by abstracting away the low-level object manipulation problem. The internal state contains the (x,y,z)-coordinates of the cursor.

\paragraph{Sawyer and Baxter}
The Sawyer and Baxter robots are available for simulation in the environment. Sawyer uses 7 DoF robotic arm, while the Baxter robot has two 7 DoF arms. They can be controlled either through impedance control and task space control (end-effector pose). For task space control, we use the PyBullet library~\citep{coumans2015bullet} to calculate inverse kinematics for the end-effector control. Unlike the cursor agent, object selection is not abstracted, and objects need to be grasped realistically by the grippers using contact forces. The internal state contains the joint angles and gripping status of the robot. An egocentric or third-person viewpoint can be selected for the camera. RGB, segmentation, and depth modes are available for the camera.

\Skip{
\begin{table}[h]
  \caption{Environment details}
  \label{table:env_detail}
  \centering
  \begin{tabular}{lccc}
    \toprule
    & \textsc{Cursor} & \textsc{Sawyer} & \textsc{Baxter} \\
    \midrule
    Observation Space & $3 \times 32 \times 32$ & $3 \times 64 \times 64$ & $4 \times 64 \times 64$ \\
    Action Space Dimension & 7 & 8 & 12 \\
    - Moving Space Dimension & $8 \times 8$ & $5 \times 3 \times 5$ & $8 \times 1 \times 8$ \\
    - Move Actions & 4 & 6 & 6 \\
    - Rotating Space Dimension & 4 & 0 & 26 \\
    - Rotate Actions & 1 & 0 & 4 \\
    \bottomrule
  \end{tabular}
\end{table}
}

\subsection{Domain Randomization}
\begin{figure*}[h]
\centering
    \includegraphics[width=1\textwidth]{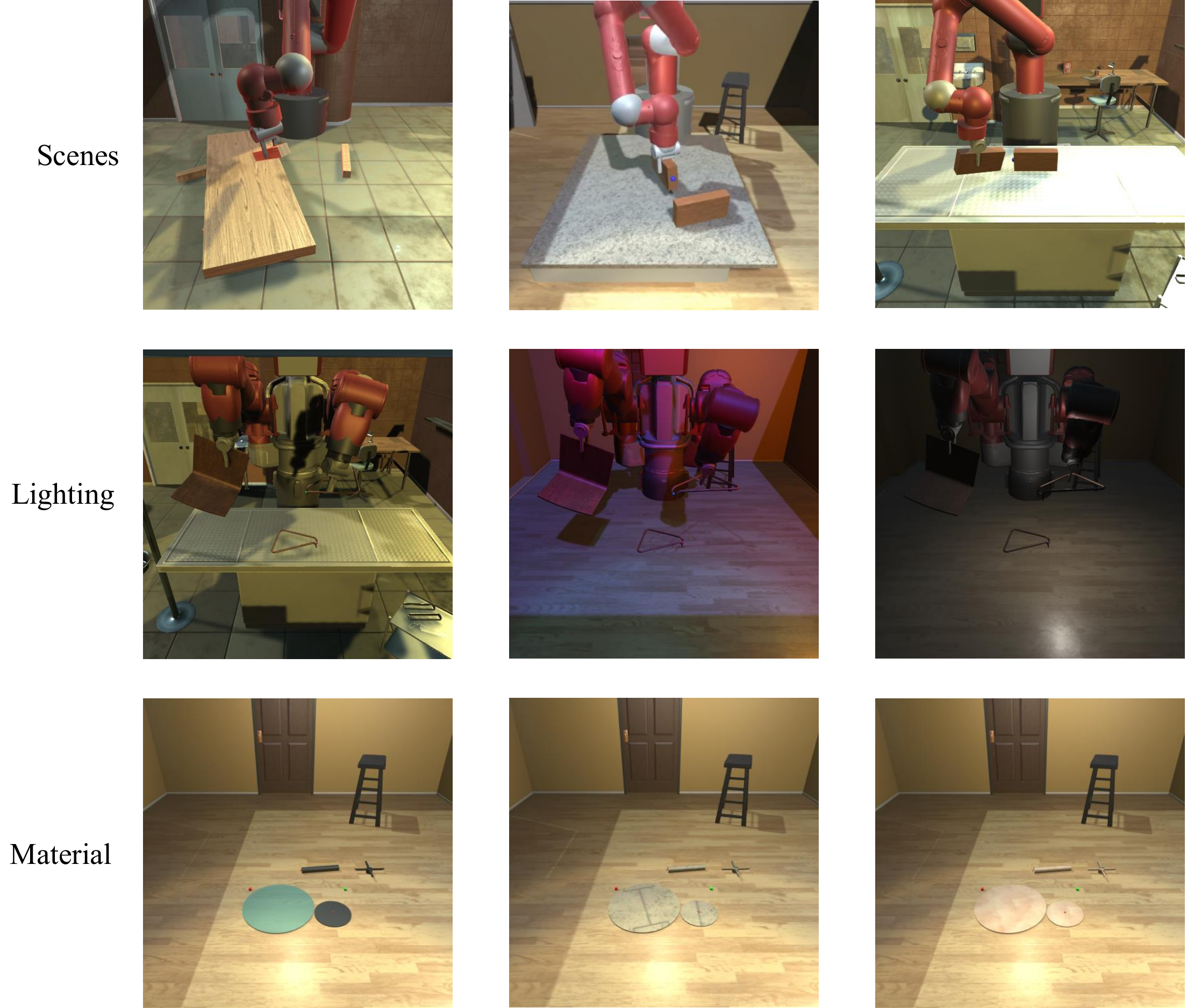}
    \caption{
        Examples of visual randomization. The first row shows different scenes such as indoors, industrial, and lab settings. The second row shows different lighting configurations like interior, low-visibility, and ambient lighting. The final row shows variations in furniture textures like plastic, marble, and sandstone.
        \label{fig:randomization}
    }
\end{figure*}

To reduce the reality gap~\citep{jakobi1995noise}, all the furniture models are created following the IKEA's official user's manuals with a minor simplification in small details such as carving and screws. 
For generalization of the learned skills, the environment should provide enough variability in furniture compositions, visual appearances, object shapes, and physical properties. For example, 
the environment will contain a diverse set of furniture including chair, table, cabinet, bookcase, desk, shelf, and tv unit (see Figure~\ref{fig:models}). For given furniture, the environment can randomly spawn subsets of the furniture parts, and randomly initialize their positions and orientation them in the scene to increase generalization. In addition to random part selection and placement, the environment can also randomize physics such as gravity and friction to add more variation in the task. The environment will also support the randomization of visual properties like lighting, background colors, texture, and more. \figref{fig:randomization} illustrates examples of domain randomization. 

\section{Limitations and Future Work}

Our environment contains several limitations. 
First, connection process of two furniture parts are not physically simulated due to the difficulty of physics simulation of screwing. Instead of simulating screws, the environment allows two parts to be connected if two connectors of parts are close and well-aligned as described in Equation~\eqref{eq:attachable}.
Next, identical furniture pieces are considered differently and one connector can only be connected to the predefined furniture part, unlike the real world where the identical pieces can be used interchangeably. In the next release, we plan to allow a furniture part replacing another identical part.

The followings are possible future extensions:
\begin{itemize}
    \item \textbf{3D motion devices:} VR controllers and 3D mouses can be used to teleoperate robots to collect human demonstration. The data can be used for training robot agents.
    \item \textbf{Additional robot support:} more diverse robots, such as Fetch, UR, and higher DoF grippers, can be trained using the proposed environment.
    \item \textbf{Realistic part attachment:} the connection mechanism can be more realistic by implementing peg insertion, screwing, nailing mechanisms.
    \item \textbf{Tool use:} an agent can learn to use a screwdriver or a hammer for assembling furniture.
    \item \textbf{Multi-agent assembly:} multiple robots collaborate to assemble complex furniture.
    \item \textbf{Instructions and demonstrations:} language and visual instructions or demonstrations can be used to guide an agent to solve a complex manipulation task.
\end{itemize}
\section{Related Work}

\subsection{Simulated Environments for Reinforcement Learning and Robot Learning}
Reinforcement learning has made rapid progress with the advent of standardized and simulated environments.
Most progress has been made in game environments, such as Atari~\citep{bellemare13arcade}, OpenAI gym~\citep{brockman2016openai}, VizDoom~\citep{Kempka2016ViZDoom}, and StarCraft2~\citep{vinyals2017starcraft}.
Recently, many simulated environments have been introduced in diverse applications, such as autonomous driving~\citep{shah2017airsim,dosovitskiy17carla}, indoor navigation~\citep{ai2thor, puig2018virtualhome}, continuous control~\citep{tassa2018deepmind, DNC, lee2019composing}, and recommendation systems~\citep{rohde2018recogym}.

In robotics, most existing environments focus on short-term object manipulation tasks, such as picking and placing~\citep{zhu2018reinforcement,lee2019silo}, in-hand dexterous manipulation~\citep{andrychowicz2018learning, rajeswaran2018learning}, door opening~\citep{urakami2019doorgym}, and peg inserting~\citep{chebotar2018closing}.
Recent advancements in simulators have made a push towards more complex and realistic tasks. \cite{fan2018surreal} takes a step towards a comprehensive manipulation simulator by offering a variety of manipulation tasks. However, the tasks, consisting of lifting, stacking, pick and place, are still limited to primitive skills. 

Composite manipulation tasks such as block stacking~\citep{duan2017one-shot, xu2018NTP} and ball serving~\citep{lee2019composing} have also been proposed. However, these tasks have small variations in shapes and physical properties of objects. Metaworld and RLBench ~\citep{yu2019meta, james2019rlbench} offer environments with diverse but simple manipulation tasks. 
In contrast, we propose a complex manipulation task with a hierarchical task structure and long horizon, \textit{furniture assembly}, which requires long-term planning and generalizable skills for various shapes, textures, and materials of objects.

\subsection{Domain Randomization}
The bottleneck of transferring an agent trained in simulation to real-world robots is the reality gap between simulation and the real world~\citep{jakobi1995noise}. 
Recently, domain randomization technique~\citep{sadeghi2017cad2rl} has been proposed to reduce the reality gap by training the policy on multiple instances of data differing in textures, backgrounds, lighting and more. Once trained with enough variation, the policy should be able to generalize to the real world~\citep{rajeswaran2016epopt, tobin2017domain,sadeghi2017sim2real, andrychowicz2018learning, urakami2019doorgym}.
To enable transfer learning of the challenging furniture assembly task, the IKEA Furniture Assembly Environment supports diverse and configurable textures, backgrounds, and lighting. Moreover, we provide over 80 furniture models with 3 different robots for further domain randomization.

\subsection{Furniture Assembly Datasets}
Solving the furniture assembly task requires the solving of multiple subproblems, such as perception, planning, and control. Each of these sub-problems are nontrivial and requires domain knowledge and supervision.
For recognizing furniture models and their structures, IKEA 3D model dataset~\citep{lim2013parsing} provide labeled data about 3D structures of furniture models and part information.
The IKEA furniture assembly video dataset~\citep{han2017human}, which consists of 480,000 video frames of humans assembling a table from IKEA, can be used as demonstration data for imitation learning.
Our proposed environment can be also used to generate synthetic training data for a wide range of 3D perception models with instance segmentation masks, depth image, and furniture part information under diverse furniture models and backgrounds.
The expert demonstration videos and trajectories can be collected and used for imitation learning, similar to Roboturk~\citep{mandlekar2018roboturk}.

\subsection{Physics Simulation Frameworks}
Mujoco~\citep{todorov2012mujoco}, Unity ML-Agents~\citep{juliani2018unity}, PyBullet~\citep{coumans2015bullet}, and Dart~\citep{lee2018dart} are some popular frameworks for physics simulation for reinforcement learning. We chose to use MuJoCo due to fast simulation and its widespread usage in the reinforcement learning community. MuJoCo also supports a Unity plugin for rendering MuJoCo scenes, enabling realistic lighting, scene, and background variations for domain randomization~\citep{sadeghi2017cad2rl}.

\section{Conclusion}

In this paper, we propose the IKEA Furniture Assembly Environment as a novel benchmark for testing complex manipulation tasks. Furniture assembly is a complex manipulation task even for humans, requiring perception capabilities, high-level planning, and sophisticated low-level control. Therefore, it is well suited as a benchmark for control algorithms aiming to solve complex tasks. 

We present several directions for future research that are necessary for completely automating furniture assembly.

\textbf{Representation}: Currently, many manipulation methods still require the pose of the object. This is difficult to obtain in the real world. On the other hand, images of the scene are easy to get but obtaining relevant features is difficult due to its high dimensionality. To solve furniture assembly in the real world, extracting object-centric representations from images is a promising direction. These representations may have geometric, relational, and semantic information~\citep{hu2018Relation, zambaldi2018deep, greff2017Neural}. Unsupervised object discovery with deep neural networks~\citep{greff2017Neural, burgess2019monet} can be also helpful.

\textbf{Planning}:
While reinforcement learning has seen success in manipulation tasks, the manipulation tasks are short in horizon and do not require complex planning. Furniture assembly on the other hand, requires long-term planning such as deciding the ordering of parts to assemble. Hierarchical reinforcement learning can segment complex tasks into simpler subtasks, which may be conducive towards furniture assembly~\citep{sutton1999option, lee2019composing}. Model-based reinforcement learning could be another direction for tackling the long-horizon problem by giving the model a mechanism to predict the future before taking an action~\citep{sutton1991dyna}.

\textbf{Dexterous Control}:
Dexterous manipulation is challenging due to agent and object complexity. Real world objects have irregular shapes, textures, friction, and other physical characteristics. Some dexterous robotic hands have between 24-30 degrees of freedom. Model free methods have seen success in controlling such dexterous hands \citep{andrychowicz2017hindsight, rajeswaran2018learning} but gaining full control of robotic hands and grasping arbitrary objects is an open challenge. Multi-agent manipulation of furniture is also another promising direction.

\textbf{Domain Knowledge}:
Instruction manuals~\citep{andreas2017modular, oh2017zero-shot}, programs~\citep{xu2018NTP, sun2018neural}, and video demonstrations~\citep{lee2019silo} can be used as domain knowledge to help solving complex tasks. The question of how to integrate additional supervision into an automated furniture assembly process is an interesting future direction.

\subsubsection*{Acknowledgments}
The authors would like to thank Taehoon Kim and many members of the USC CLVR lab for helpful feedback and testing our environment. This project was partially funded by SKT.

\clearpage

\bibliography{bib/drl,bib/detection,bib/deep_learning,bib/rl,bib/hrl,bib/env,bib/gnn,bib/sim2real,bib/robotics,bib/goal_directed_rl,bib/meta,bib/imitation,bib/datasets}

\begin{thebibliography}{46}
\providecommand{\natexlab}[1]{#1}
\providecommand{\url}[1]{\texttt{#1}}
\expandafter\ifx\csname urlstyle\endcsname\relax
  \providecommand{\doi}[1]{doi: #1}\else
  \providecommand{\doi}{doi: \begingroup \urlstyle{rm}\Url}\fi

\bibitem[Oh et~al.(2017)Oh, Singh, Lee, and Kohli]{oh2017zero-shot}
Junhyuk Oh, Satinder Singh, Honglak Lee, and Pushmeet Kohli.
\newblock Zero-shot task generalization with multi-task deep reinforcement
  learning.
\newblock In \emph{International Conference on Learning Representations}, pages
  2661--2670, 2017.

\bibitem[Chaplot et~al.(2017)Chaplot, Sathyendra, Pasumarthi, Rajagopal, and
  Salakhutdinov]{chaplot2017gated}
Devendra~Singh Chaplot, Kanthashree~Mysore Sathyendra, Rama~Kumar Pasumarthi,
  Dheeraj Rajagopal, and Ruslan Salakhutdinov.
\newblock Gated-attention architectures for task-oriented language grounding.
\newblock \emph{arXiv preprint arXiv:1706.07230}, 2017.

\bibitem[Todorov et~al.(2012)Todorov, Erez, and Tassa]{todorov2012mujoco}
Emanuel Todorov, Tom Erez, and Yuval Tassa.
\newblock Mujoco: A physics engine for model-based control.
\newblock In \emph{IEEE/RSJ International Conference on Intelligent Robots and
  Systems}, pages 5026--5033, 2012.

\bibitem[Fan et~al.(2018)Fan, Zhu, Zhu, Liu, Zeng, Gupta, Creus-Costa,
  Savarese, and Fei-Fei]{fan2018surreal}
Linxi Fan, Yuke Zhu, Jiren Zhu, Zihua Liu, Orien Zeng, Anchit Gupta, Joan
  Creus-Costa, Silvio Savarese, and Li~Fei-Fei.
\newblock Surreal: Open-source reinforcement learning framework and robot
  manipulation benchmark.
\newblock In \emph{Conference on Robot Learning}, pages 767--782, 2018.

\bibitem[Urakami et~al.(2019)Urakami, Hodgkinson, Carlin, Leu, Rigazio, and
  Abbeel]{urakami2019doorgym}
Yusuke Urakami, Alec Hodgkinson, Casey Carlin, Randall Leu, Luca Rigazio, and
  Pieter Abbeel.
\newblock Doorgym: A scalable door opening environment and baseline agent.
\newblock \emph{arXiv preprint arXiv:1908.01887}, 2019.

\bibitem[Brockman et~al.(2016)Brockman, Cheung, Pettersson, Schneider,
  Schulman, Tang, and Zaremba]{brockman2016openai}
Greg Brockman, Vicki Cheung, Ludwig Pettersson, Jonas Schneider, John Schulman,
  Jie Tang, and Wojciech Zaremba.
\newblock Openai gym.
\newblock \emph{arXiv preprint arXiv:1606.01540}, 2016.

\bibitem[Coumans(2015)]{coumans2015bullet}
Erwin Coumans.
\newblock Bullet physics simulation.
\newblock In \emph{ACM SIGGRAPH 2015 Courses}, SIGGRAPH '15, New York, NY, USA,
  2015. ACM.
\newblock ISBN 978-1-4503-3634-5.
\newblock \doi{10.1145/2776880.2792704}.
\newblock URL \url{http://doi.acm.org/10.1145/2776880.2792704}.

\bibitem[Jakobi et~al.(1995)Jakobi, Husbands, and Harvey]{jakobi1995noise}
Nick Jakobi, Phil Husbands, and Inman Harvey.
\newblock Noise and the reality gap: The use of simulation in evolutionary
  robotics.
\newblock In \emph{European Conference on Artificial Life}, pages 704--720.
  Springer, 1995.

\bibitem[{Bellemare} et~al.(2013){Bellemare}, {Naddaf}, {Veness}, and
  {Bowling}]{bellemare13arcade}
M.~G. {Bellemare}, Y.~{Naddaf}, J.~{Veness}, and M.~{Bowling}.
\newblock The arcade learning environment: An evaluation platform for general
  agents.
\newblock \emph{Journal of Artificial Intelligence Research}, 47:\penalty0
  253--279, jun 2013.

\bibitem[Kempka et~al.(2016)Kempka, Wydmuch, Runc, Toczek, and
  Ja\'skowski]{Kempka2016ViZDoom}
Micha{\l} Kempka, Marek Wydmuch, Grzegorz Runc, Jakub Toczek, and Wojciech
  Ja\'skowski.
\newblock {ViZDoom}: A {D}oom-based {AI} research platform for visual
  reinforcement learning.
\newblock In \emph{IEEE Conference on Computational Intelligence and Games},
  pages 341--348, 2016.

\bibitem[Vinyals et~al.(2017)Vinyals, Ewalds, Bartunov, Georgiev, Vezhnevets,
  Yeo, Makhzani, K{\"u}ttler, Agapiou, Schrittwieser,
  et~al.]{vinyals2017starcraft}
Oriol Vinyals, Timo Ewalds, Sergey Bartunov, Petko Georgiev, Alexander~Sasha
  Vezhnevets, Michelle Yeo, Alireza Makhzani, Heinrich K{\"u}ttler, John
  Agapiou, Julian Schrittwieser, et~al.
\newblock Starcraft ii: A new challenge for reinforcement learning.
\newblock \emph{arXiv preprint arXiv:1708.04782}, 2017.

\bibitem[Shah et~al.(2017)Shah, Dey, Lovett, and Kapoor]{shah2017airsim}
Shital Shah, Debadeepta Dey, Chris Lovett, and Ashish Kapoor.
\newblock Airsim: High-fidelity visual and physical simulation for autonomous
  vehicles.
\newblock In \emph{Field and Service Robotics}, 2017.
\newblock URL \url{https://arxiv.org/abs/1705.05065}.

\bibitem[Dosovitskiy et~al.(2017)Dosovitskiy, Ros, Codevilla, Lopez, and
  Koltun]{dosovitskiy17carla}
Alexey Dosovitskiy, German Ros, Felipe Codevilla, Antonio Lopez, and Vladlen
  Koltun.
\newblock {CARLA}: {An} open urban driving simulator.
\newblock In \emph{Conference on Robot Learning}, pages 1--16, 2017.

\bibitem[Kolve et~al.(2017)Kolve, Mottaghi, Gordon, Zhu, Gupta, and
  Farhadi]{ai2thor}
Eric Kolve, Roozbeh Mottaghi, Daniel Gordon, Yuke Zhu, Abhinav Gupta, and Ali
  Farhadi.
\newblock Ai2-thor: An interactive 3d environment for visual ai.
\newblock \emph{arXiv preprint arXiv:1712.05474}, 2017.

\bibitem[Puig et~al.(2018)Puig, Ra, Boben, Li, Wang, Fidler, and
  Torralba]{puig2018virtualhome}
Xavier Puig, Kevin Ra, Marko Boben, Jiaman Li, Tingwu Wang, Sanja Fidler, and
  Antonio Torralba.
\newblock Virtualhome: Simulating household activities via programs.
\newblock In \emph{Proceedings of the IEEE Conference on Computer Vision and
  Pattern Recognition}, pages 8494--8502, 2018.

\bibitem[Tassa et~al.(2018)Tassa, Doron, Muldal, Erez, Li, Casas, Budden,
  Abdolmaleki, Merel, Lefrancq, et~al.]{tassa2018deepmind}
Yuval Tassa, Yotam Doron, Alistair Muldal, Tom Erez, Yazhe Li, Diego de~Las
  Casas, David Budden, Abbas Abdolmaleki, Josh Merel, Andrew Lefrancq, et~al.
\newblock Deepmind control suite.
\newblock \emph{arXiv preprint arXiv:1801.00690}, 2018.

\bibitem[Ghosh et~al.(2018)Ghosh, Singh, Rajeswaran, Kumar, and Levine]{DNC}
Dibya Ghosh, Avi Singh, Aravind Rajeswaran, Vikash Kumar, and Sergey Levine.
\newblock Divide and conquer reinforcement learning.
\newblock In \emph{International Conference on Learning Representations}, 2018.

\bibitem[Lee et~al.(2019{\natexlab{a}})Lee, Sun, Somasundaram, Hu, and
  Lim]{lee2019composing}
Youngwoon Lee, Shao-Hua Sun, Sriram Somasundaram, Edward~S. Hu, and Joseph~J.
  Lim.
\newblock Composing complex skills by learning transition policies.
\newblock In \emph{International Conference on Learning Representations},
  2019{\natexlab{a}}.
\newblock URL \url{https://openreview.net/forum?id=rygrBhC5tQ}.

\bibitem[Rohde et~al.(2018)Rohde, Bonner, Dunlop, Vasile, and
  Karatzoglou]{rohde2018recogym}
David Rohde, Stephen Bonner, Travis Dunlop, Flavian Vasile, and Alexandros
  Karatzoglou.
\newblock Recogym: A reinforcement learning environment for the problem of
  product recommendation in online advertising.
\newblock \emph{arXiv preprint arXiv:1808.00720}, 2018.

\bibitem[Zhu et~al.(2018)Zhu, Wang, Merel, Rusu, Erez, Cabi, Tunyasuvunakool,
  Kram{\'a}r, Hadsell, de~Freitas, and Heess]{zhu2018reinforcement}
Yuke Zhu, Ziyu Wang, Josh Merel, Andrei Rusu, Tom Erez, Serkan Cabi, Saran
  Tunyasuvunakool, J{\'a}nos Kram{\'a}r, Raia Hadsell, Nando de~Freitas, and
  Nicolas Heess.
\newblock Reinforcement and imitation learning for diverse visuomotor skills.
\newblock \emph{arXiv preprint arXiv:1802.09564}, 2018.

\bibitem[Lee et~al.(2019{\natexlab{b}})Lee, Hu, Yang, and Lim]{lee2019silo}
Youngwoon Lee, Edward~S. Hu, Zhengyu Yang, and Joseph~J. Lim.
\newblock To follow or not to follow: Selective imitation learning from
  observations.
\newblock In \emph{Conference on Robot Learning}, 2019{\natexlab{b}}.

\bibitem[Andrychowicz et~al.(2018)Andrychowicz, Baker, Chociej, Jozefowicz,
  McGrew, Pachocki, Petron, Plappert, Powell, Ray,
  et~al.]{andrychowicz2018learning}
Marcin Andrychowicz, Bowen Baker, Maciek Chociej, Rafal Jozefowicz, Bob McGrew,
  Jakub Pachocki, Arthur Petron, Matthias Plappert, Glenn Powell, Alex Ray,
  et~al.
\newblock Learning dexterous in-hand manipulation.
\newblock \emph{arXiv preprint arXiv:1808.00177}, 2018.

\bibitem[Rajeswaran et~al.(2018)Rajeswaran, Kumar, Gupta, Vezzani, Schulman,
  Todorov, and Levine]{rajeswaran2018learning}
Aravind Rajeswaran, Vikash Kumar, Abhishek Gupta, Giulia Vezzani, John
  Schulman, Emanuel Todorov, and Sergey Levine.
\newblock Learning complex dexterous manipulation with deep reinforcement
  learning and demonstrations.
\newblock In \emph{Robotics: Science and Systems}, 2018.

\bibitem[Chebotar et~al.(2018)Chebotar, Handa, Makoviychuk, Macklin, Issac,
  Ratliff, and Fox]{chebotar2018closing}
Yevgen Chebotar, Ankur Handa, Viktor Makoviychuk, Miles Macklin, Jan Issac,
  Nathan Ratliff, and Dieter Fox.
\newblock Closing the sim-to-real loop: Adapting simulation randomization with
  real world experience.
\newblock \emph{arXiv preprint arXiv:1810.05687}, 2018.

\bibitem[Duan et~al.(2017)Duan, Andrychowicz, Stadie, Ho, Schneider, Sutskever,
  Abbeel, and Zaremba]{duan2017one-shot}
Yan Duan, Marcin Andrychowicz, Bradly Stadie, Jonathan Ho, Jonas Schneider,
  Ilya Sutskever, Pieter Abbeel, and Wojciech Zaremba.
\newblock One-shot imitation learning.
\newblock In \emph{Advances in Neural Information Processing Systems}, pages
  1087--1098, 2017.

\bibitem[Xu et~al.(2018)Xu, Nair, Zhu, Gao, Garg, Fei-Fei, and
  Savarese]{xu2018NTP}
Danfei Xu, Suraj Nair, Yuke Zhu, Julian Gao, Animesh Garg, Li~Fei-Fei, and
  Silvio Savarese.
\newblock Neural task programming: Learning to generalize across hierarchical
  tasks.
\newblock In \emph{IEEE International Conference on Robotics and Automation},
  pages 1--8. IEEE, 2018.

\bibitem[Yu et~al.(2019)Yu, Quillen, He, Julian, Hausman, Finn, and
  Levine]{yu2019meta}
Tianhe Yu, Deirdre Quillen, Zhanpeng He, Ryan Julian, Karol Hausman, Chelsea
  Finn, and Sergey Levine.
\newblock Meta-world: A benchmark and evaluation for multi-task and meta
  reinforcement learning.
\newblock In \emph{Conference on Robot Learning (CoRL)}, 2019.

\bibitem[James et~al.(2019)James, Ma, Rovick~Arrojo, and
  Davison]{james2019rlbench}
Stephen James, Zicong Ma, David Rovick~Arrojo, and Andrew~J. Davison.
\newblock Rlbench: The robot learning benchmark \& learning environment.
\newblock \emph{arXiv preprint arXiv:1909.12271}, 2019.

\bibitem[Sadeghi and Levine(2017)]{sadeghi2017cad2rl}
Fereshteh Sadeghi and Sergey Levine.
\newblock Cad2rl: Real single-image flight without a single real image.
\newblock In \emph{Robotics: Science and Systems}, 2017.

\bibitem[Rajeswaran et~al.(2016)Rajeswaran, Ghotra, Ravindran, and
  Levine]{rajeswaran2016epopt}
Aravind Rajeswaran, Sarvjeet Ghotra, Balaraman Ravindran, and Sergey Levine.
\newblock Epopt: Learning robust neural network policies using model ensembles.
\newblock \emph{arXiv preprint arXiv:1610.01283}, 2016.

\bibitem[Tobin et~al.(2017)Tobin, Fong, Ray, Schneider, Zaremba, and
  Abbeel]{tobin2017domain}
Josh Tobin, Rachel Fong, Alex Ray, Jonas Schneider, Wojciech Zaremba, and
  Pieter Abbeel.
\newblock Domain randomization for transferring deep neural networks from
  simulation to the real world.
\newblock In \emph{IEEE/RSJ International Conference on Intelligent Robots and
  Systems}, pages 23--30. IEEE, 2017.

\bibitem[Sadeghi et~al.(2017)Sadeghi, Toshev, Jang, and
  Levine]{sadeghi2017sim2real}
Fereshteh Sadeghi, Alexander Toshev, Eric Jang, and Sergey Levine.
\newblock Sim2real view invariant visual servoing by recurrent control.
\newblock \emph{arXiv preprint arXiv:1712.07642}, 2017.

\bibitem[Lim et~al.(2013)Lim, Pirsiavash, and Torralba]{lim2013parsing}
Joseph~J Lim, Hamed Pirsiavash, and Antonio Torralba.
\newblock Parsing ikea objects: Fine pose estimation.
\newblock In \emph{IEEE Conference on Computer Vision and Pattern Recognition},
  pages 2992--2999, 2013.

\bibitem[Han et~al.(2017)Han, Wang, Cherian, and Gould]{han2017human}
Tengda Han, Jue Wang, Anoop Cherian, and Stephen Gould.
\newblock Human action forecasting by learning task grammars.
\newblock \emph{arXiv preprint arXiv:1709.06391}, 2017.

\bibitem[Mandlekar et~al.(2018)Mandlekar, Zhu, Garg, Booher, Spero, Tung, Gao,
  Emmons, Gupta, Orbay, Savarese, and Fei{-}Fei]{mandlekar2018roboturk}
Ajay Mandlekar, Yuke Zhu, Animesh Garg, Jonathan Booher, Max Spero, Albert
  Tung, Julian Gao, John Emmons, Anchit Gupta, Emre Orbay, Silvio Savarese, and
  Li~Fei{-}Fei.
\newblock Roboturk: {A} crowdsourcing platform for robotic skill learning
  through imitation.
\newblock \emph{CoRR}, abs/1811.02790, 2018.
\newblock URL \url{http://arxiv.org/abs/1811.02790}.

\bibitem[Juliani et~al.(2018)Juliani, Berges, Vckay, Gao, Henry, Mattar, and
  Lange]{juliani2018unity}
Arthur Juliani, Vincent-Pierre Berges, Esh Vckay, Yuan Gao, Hunter Henry,
  Marwan Mattar, and Danny Lange.
\newblock Unity: A general platform for intelligent agents.
\newblock \emph{arXiv preprint arXiv:1809.02627}, 2018.

\bibitem[Lee et~al.(2018)Lee, Grey, Ha, Kunz, Jain, Ye, Srinivasa, Stilman, and
  Liu]{lee2018dart}
Jeongseok Lee, Michael~X Grey, Sehoon Ha, Tobias Kunz, Sumit Jain, Yuting Ye,
  Siddhartha~S Srinivasa, Mike Stilman, and C~Karen Liu.
\newblock Dart: Dynamic animation and robotics toolkit.
\newblock \emph{The Journal of Open Source Software}, 3\penalty0 (22):\penalty0
  500, 2018.

\bibitem[Hu et~al.(2018)Hu, Gu, Zhang, Dai, and Wei]{hu2018Relation}
Han Hu, Jiayuan Gu, Zheng Zhang, Jifeng Dai, and Yichen Wei.
\newblock Relation networks for object detection.
\newblock In \emph{IEEE Conference on Computer Vision and Pattern Recognition},
  pages 3588--3597, 2018.

\bibitem[Zambaldi et~al.(2019)Zambaldi, Raposo, Santoro, Bapst, Li, Babuschkin,
  Tuyls, Reichert, Lillicrap, Lockhart, Shanahan, Langston, Pascanu, Botvinick,
  Vinyals, and Battaglia]{zambaldi2018deep}
Vinicius Zambaldi, David Raposo, Adam Santoro, Victor Bapst, Yujia Li, Igor
  Babuschkin, Karl Tuyls, David Reichert, Timothy Lillicrap, Edward Lockhart,
  Murray Shanahan, Victoria Langston, Razvan Pascanu, Matthew Botvinick, Oriol
  Vinyals, and Peter Battaglia.
\newblock Deep reinforcement learning with relational inductive biases.
\newblock In \emph{International Conference on Learning Representations}, 2019.
\newblock URL \url{https://openreview.net/forum?id=HkxaFoC9KQ}.

\bibitem[Greff et~al.(2017)Greff, van Steenkiste, and
  Schmidhuber]{greff2017Neural}
Klaus Greff, Sjoerd van Steenkiste, and J\"{u}rgen Schmidhuber.
\newblock Neural expectation maximization.
\newblock In I.~Guyon, U.~V. Luxburg, S.~Bengio, H.~Wallach, R.~Fergus,
  S.~Vishwanathan, and R.~Garnett, editors, \emph{Advances in Neural
  Information Processing Systems}, pages 6691--6701, 2017.

\bibitem[Burgess et~al.(2019)Burgess, Matthey, Watters, Kabra, Higgins,
  Botvinick, and Lerchner]{burgess2019monet}
Christopher~P. Burgess, Lo{\"{\i}}c Matthey, Nicholas Watters, Rishabh Kabra,
  Irina Higgins, Matthew Botvinick, and Alexander Lerchner.
\newblock Monet: Unsupervised scene decomposition and representation.
\newblock \emph{CoRR}, abs/1901.11390, 2019.
\newblock URL \url{http://arxiv.org/abs/1901.11390}.

\bibitem[Sutton et~al.(1999)Sutton, Precup, and Singh]{sutton1999option}
Richard~S Sutton, Doina Precup, and Satinder Singh.
\newblock Between mdps and semi-mdps: A framework for temporal abstraction in
  reinforcement learning.
\newblock \emph{Artificial intelligence}, 112\penalty0 (1-2):\penalty0
  181--211, 1999.

\bibitem[Sutton(1991)]{sutton1991dyna}
Richard~S Sutton.
\newblock Dyna, an integrated architecture for learning, planning, and
  reacting.
\newblock \emph{ACM Sigart Bulletin}, 2\penalty0 (4):\penalty0 160--163, 1991.

\bibitem[Andrychowicz et~al.(2017)Andrychowicz, Wolski, Ray, Schneider, Fong,
  Welinder, McGrew, Tobin, Abbeel, and Zaremba]{andrychowicz2017hindsight}
Marcin Andrychowicz, Filip Wolski, Alex Ray, Jonas Schneider, Rachel Fong,
  Peter Welinder, Bob McGrew, Josh Tobin, Pieter Abbeel, and Wojciech Zaremba.
\newblock Hindsight experience replay.
\newblock In \emph{Advances in Neural Information Processing Systems}, pages
  5048--5058, 2017.

\bibitem[Andreas et~al.(2017)Andreas, Klein, and Levine]{andreas2017modular}
Jacob Andreas, Dan Klein, and Sergey Levine.
\newblock Modular multitask reinforcement learning with policy sketches.
\newblock In \emph{International Conference on Machine Learning}, pages
  166--175, 2017.

\bibitem[Sun et~al.(2018)Sun, Noh, Somasundaram, and Lim]{sun2018neural}
Shao-Hua Sun, Hyeonwoo Noh, Sriram Somasundaram, and Joseph~J. Lim.
\newblock Neural program synthesis from diverse demonstration videos.
\newblock In \emph{International Conference on Machine Learning}, 2018.

\end{thebibliography}
\bibliographystyle{unsrtnat}

\end{document}